\documentclass[10pt,twocolumn,letterpaper]{article}

\usepackage{iccv}              

\usepackage{booktabs}
\usepackage{multirow}
\usepackage{algorithm}
\usepackage{algpseudocode}
\usepackage{marvosym}
\usepackage{graphicx}
\usepackage{pifont}
\usepackage[algo2e,linesnumbered,noend]{algorithm2e}

\definecolor{iccvblue}{rgb}{0.21,0.49,0.74}
\usepackage[pagebackref,breaklinks,colorlinks,allcolors=iccvblue]{hyperref}

\def\paperID{10550} 
\def\confName{ICCV}
\def\confYear{2025}

\title{Differential-informed Sample Selection Accelerates Multimodal Contrastive Learning}

\author{Zihua Zhao$^1$*, Feng Hong$^1$*, Mengxi Chen$^{1}$, Pengyi Chen$^{1}$, Benyuan Liu$^{1}$, Jiangchao Yao$\textsuperscript{1,3\ \Letter}$, \\ Ya Zhang$^{2,3}$, Yanfeng Wang$\textsuperscript{2,3\ \Letter}$\\
$^1$ Cooperative Medianet Innovation Center, Shanghai Jiao Tong University \\ $^2$ School of AI, Shanghai Jiao Tong University $^3$ Shanghai AI Laboratory\\
{\tt\footnotesize \{sjtuszzh, feng.hong, mxchen\_mc, chenmuqiu, lby0912, Sunarker, ya\_zhang, wangyanfeng\}@sjtu.edu.cn}
}

\begin{document}
\maketitle

\renewcommand\thefootnote{*}
\footnotetext{The first two authors contribute equally.}

\begin{abstract}

The remarkable success of contrastive-learning-based multimodal models has been greatly driven by training on ever-larger datasets with expensive compute consumption. Sample selection as an alternative efficient paradigm plays an important direction to accelerate the training process. However, recent advances on sample selection either mostly rely on an oracle model to offline select a high-quality coreset, which is limited in the cold-start scenarios, or focus on online selection based on real-time model predictions, which has not sufficiently or efficiently considered the noisy correspondence. To address this dilemma, we propose a novel \underline{D}ifferential-\underline{I}nformed \underline{S}ample \underline{S}el\underline{ect}ion (DISSect) method, which accurately and efficiently discriminates the noisy correspondence for training acceleration. Specifically, we rethink the impact of noisy correspondence on contrastive learning and propose that the differential between the predicted correlation of the current model and that of a historical model is more informative to characterize sample quality. Based on this, we construct a robust differential-based sample selection and analyze its theoretical insights. Extensive experiments on three benchmark datasets and various downstream tasks demonstrate the consistent superiority of DISSect over current state-of-the-art methods. Source code is available at: \url{https://github.com/MediaBrain-SJTU/DISSect}.

\end{abstract}
\section{Introduction}

Multimodal models~\cite{li2022blip, cherti2023reproducible, jia2021scaling, kim2021vilt} have widely adopted the contrastive learning paradigm~\citep{oord2018representation, DBLP:journals/ml/ZhouYHZYZTW25} to capture the mutual information across modalities. While achieving remarkable success over the past years, much of this progress has been driven by training on ever-expanding datasets~\citep{sharifani2023machine, kaplan2020scaling}, which often comes at a substantial compute cost. To reduce the burden, sample selection~\citep{raju2021accelerating, sener2017active} draws the increasing attention recently, as identifying informative and diverse samples is widely recognized as essential for accelerating the model convergence and enhancing the performance.

\begin{figure}[t]
  \centering
   \includegraphics[width=1.0\linewidth]{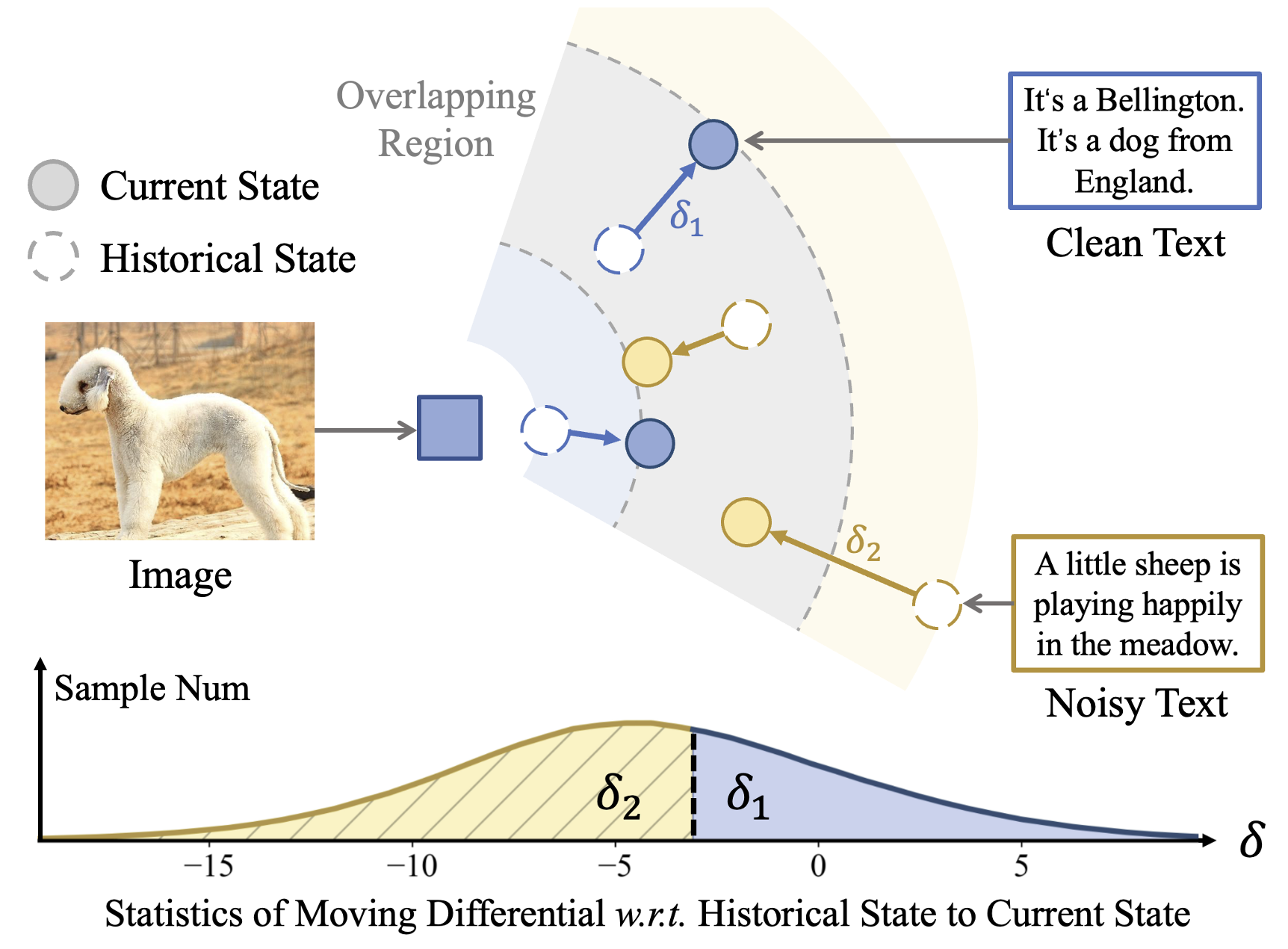}
   \caption{Learning from samples with noisy correspondence (\textit{e.g.} wrongly relate a ``Bellington" to a sheep) undermines multimodal contrastive learning. Both clean and noisy samples are memorized in an overlapping region. DISSect effectively discriminates noisy correspondence with distinct moving differential $\delta$ between hitorical and current sample learning states.}
   \label{fig:0_intro}
   \vspace{-8pt}
\end{figure}

Existing sample selection for acceleration can generally be categorized into two main lines: coreset sample selection~\citep{park2022active, zheng2022coverage} and online sample selection~\citep{he2024large, mindermann2022prioritized}. Coreset selection methods aim to build a coreset from the original data for reproducing training, with inherently an oracle model available as the auxiliary to filter out samples~\citep{hessel2021clipscore, wang2025cliploss}. The success of such methods highlights the importance of accurately eliminating noisy correspondence samples, however obtaining such an oracle model is not always feasible in some cold-start scenarios. In contrast, online selection~\citep{wu2024icons, qin2023infobatch} focuses on selecting data along with the training dynamics. However, most of these online acceleration-oriented methods have not sufficiently considered the noisy correspondence, yielding the adverse samples being  selected~\citep{hong2024diversified}. For methods that specially handle the noisy correspondence~\citep{huang2021learning, zhao2024mitigating}, they widely utilize a dual-network structure, which incurs considerable computational cost and is not efficient for acceleration.

In this paper, we aim at selecting informative samples to promote fast training of multimodal contrastive learning in an oracle-model-free online manner. As illustrated in Fig.~\ref{fig:0_intro}, we observe that under multimodal contrastive learning, the divergence of cross-modal similarity predicted by the model shrinks significantly after the initial epochs, and as training progresses, both clean pairs and correspondence noise are memorized into a close region, which cannot be easily separated by the popular loss threshold that is incorporated in previous noisy label/correspondence learning. Fortunately, the moving differential between current model prediction and the historical state seems to be more informative to characterize the sample quality according to the statistics.

Based on the above analysis, we propose a novel selection method for acceleration, namely, \underline{D}ifferential-\underline{I}nformed \underline{S}ample \underline{S}el\underline{ect}ion (DISSect). 
Specifically, we construct the sample-wise differential between the predicted correlation of the current model and that of the historical model to capture the model's learning tendency for each sample, indicating whether the sample is being forgotten or wrongly memorized. And the differential function is implemented using a discrepancy in CLIPScore~\citep{hessel2021clipscore}, and obtain the historical prediction by either initializing with a warm-up stage or maintaining a temporal ensembling update, which does not incur much cost in computation. Finally, through a ranked sampling strategy, DISSect prioritizes the training efficiently on the informative samples to achieve acceleration. Our contributions are summarized as follows:

\begin{itemize}
    \item We delve into the under-explored realm of sample selection for accelerated multimodal contrastive learning, identify key heuristics, and highlight the critical role of noise correspondence removal in this setting.
    \item We propose a novel Differential-informed Sample Selection (DISSect) approach, which utilizes the differential of similarity predictions between the historical and current model to prevent learning from noisy correspondence and accelerate multimodal contrastive learning.
    \item Extensive experiments on three benchmark datasets and various downstream tasks demonstrate the consistent superiority of DISSect over current state-of-the-art methods, achieving comparable performance to full data training with 70\% fewer iterations on the CC3M dataset.
\end{itemize}
\section{Related Work}

In this section, we discuss about the most relative works in coreset sample selection and online sample selection. Other close research directs like curriculum learning and dataset distillation are discussed in Appendix.

\subsection{Coreset sample selection}

Coreset sample selection, also known as data pruning~\cite{park2022active, zheng2022coverage} targets at creating a subset of the original data where inherent low-quality samples are filtered to reserve essential patterns for efficient model training. Previous methods~\citep{han2025trustworthy, yao2023latent} evaluate the importance of samples through proposing standardized scores, including entropy score~\citep{coleman2019selection}, EL2N score~\citep{paul2021deep} and forgetting score~\citep{toneva2018empirical}, based on which samples with high scores are selected. While some coreset selection methods have further considered the coverage and diversity of selected data~\citep{xia2022moderate, zheng2022coverage}, the limited amount of training data can inevitably degrade the model performance due to the scaling law~\citep{kaplan2020scaling}
and the sampled data can be biased for single evaluation score which neglects the demand during model training~\citep{hong2024diversified}. In the field of vision-language pre-training~\citep{chen2025multi, DBLP:journals/corr/abs-2503-22215, li2025incomplete, li2024Fast, chen2023enhanced,DBLP:journals/tmi/DaiZHYZW24}, coreset selection methods widely evaluate datasets adapting reference models. CLIPScore~\citep{hessel2021clipscore} is first proposed to evaluate pair-wise similarities between image-text data by pre-trained CLIP~\cite{radford2021learning} model. TL;DR~\citep{wang2023too} and \citet{nguyen2024improving} propose to refine the low-quality paired data through captioning by a pre-trained generative model. Sieve~\citep{mahmoud2024sieve} select better samples from correspondence quality aspect with pruning signal trained on small clean subset. Despite the better quality of the sampled coreset, these methods rely on extra reference models trained from a considerable amount of holdout data~\citep{mindermann2022prioritized}, which can be unaffordable in large scenarios.

\subsection{Online sample selection}

Online sample selection~\citep{mindermann2022prioritized, he2024large} aims at tailoring the dataset adaptive to the model status during training. The training cost is thus saved through reducing the number of iterations for convergence. InfoBatch~\citep{qin2023infobatch} randomly prunes a portion of less informative samples and rescales the gradients to approximate the original ones. DivBS~\citep{hong2024diversified} selects samples with better diversity through measuring group-wise orthogonal representativeness. ICONs~\citep{wu2024icons} utilizes majority vote across task-specific influence matrices to identify valuable samples for down-stream tasks. SCAN~\citep{guo2024scan} introduces a dynamic bootstrapping method which gradually increases pruning rate. Online sample selection has also been researched to filter out noisy correspondence samples~\citep{huang2021learning, yang2023bicro} in vision-language pre-training. NCR~\citep{huang2021learning} first arouse this issue and utilizes DivideMix~\citep{li2020dividemix} paradigm to distinguish clean pairs from noisy ones. GSC~\citep{zhao2024mitigating} further incorporates consistency of both cross-modal and intra-modal geometrical structures to recognize noisy correspondence. However, to eliminate as much noisy correspondence samples, learning from noisy correspondence methods widely adapt dual networks~\citep{han2018co}, which demand high computational cost and are not suitable for large-scale scenarios.
\section{Method}

\subsection{Preliminary}

Given a multimodal dataset $\mathcal{D} = \{I_i, T_i\}^{|\mathcal{D}|}_{i=1}$, where $\{I_i, T_i\}$ represents the $i$-th paired image-text input and $|\mathcal{D}|$ denotes the dataset size, we consider the setting under the contrastive learning paradigm proposed by CLIP~\citep{cherti2023reproducible} without losing generality~\citep{he2020momentum, li2022blip, li2023blip}. CLIP learns the mutual information across modalities by projecting image and caption inputs into representations $f(I)$ and $g(T)$ (or $I$, $T$ in brief) using separate encoders $f_{\theta_1}$ and $g_{\theta_2}$, where $\theta_1$ and $\theta_2$ are the model parameters. For alignment, CLIP closes the distance between positive pairs while contrasting them to the negative ones within the batch $\mathcal{D}_b$, using the InfoNCE loss~\citep{oord2018representation} defined as,
\begin{equation}
\begin{aligned}
 \mathcal{L}_{f\to g} =& \frac{-1}{|\mathcal{D}_b|}\sum_{i=1}^{|\mathcal{D}_b|}\log{\frac{\exp{\left(\frac{f(I_i)^\top g(T_i)}{\tau}\right)}}{\sum_{j=1}^{|\mathcal{D}_b|}\exp{\left(\frac{f(I_i)^\top g(T_j)}{\tau}\right)}}},
\label{equ:0_mthod}
\end{aligned}
\end{equation}
\noindent
where $\tau$ is a learnable temperature. The overall loss is computed as the mean of dual direction losses $\mathcal{L}_{f\to g}$ and $\mathcal{L}_{g\to f}$. Sample selection methods aim to accelerate such training procedure through tailoring the dataset. Coreset selection methods typically filter a subset $\mathcal{D'}\sim \mathcal{D}$ consisting of valued samples before training, while online selection methods dynamically select batched samples $\mathcal{D}_b'(f, g)\sim \mathcal{D}_b$ according to the training status.

\subsection{Heuristics: Multimodal Sample Selection}

\begin{figure}[t]
  \centering
   \includegraphics[width=1.0\linewidth]{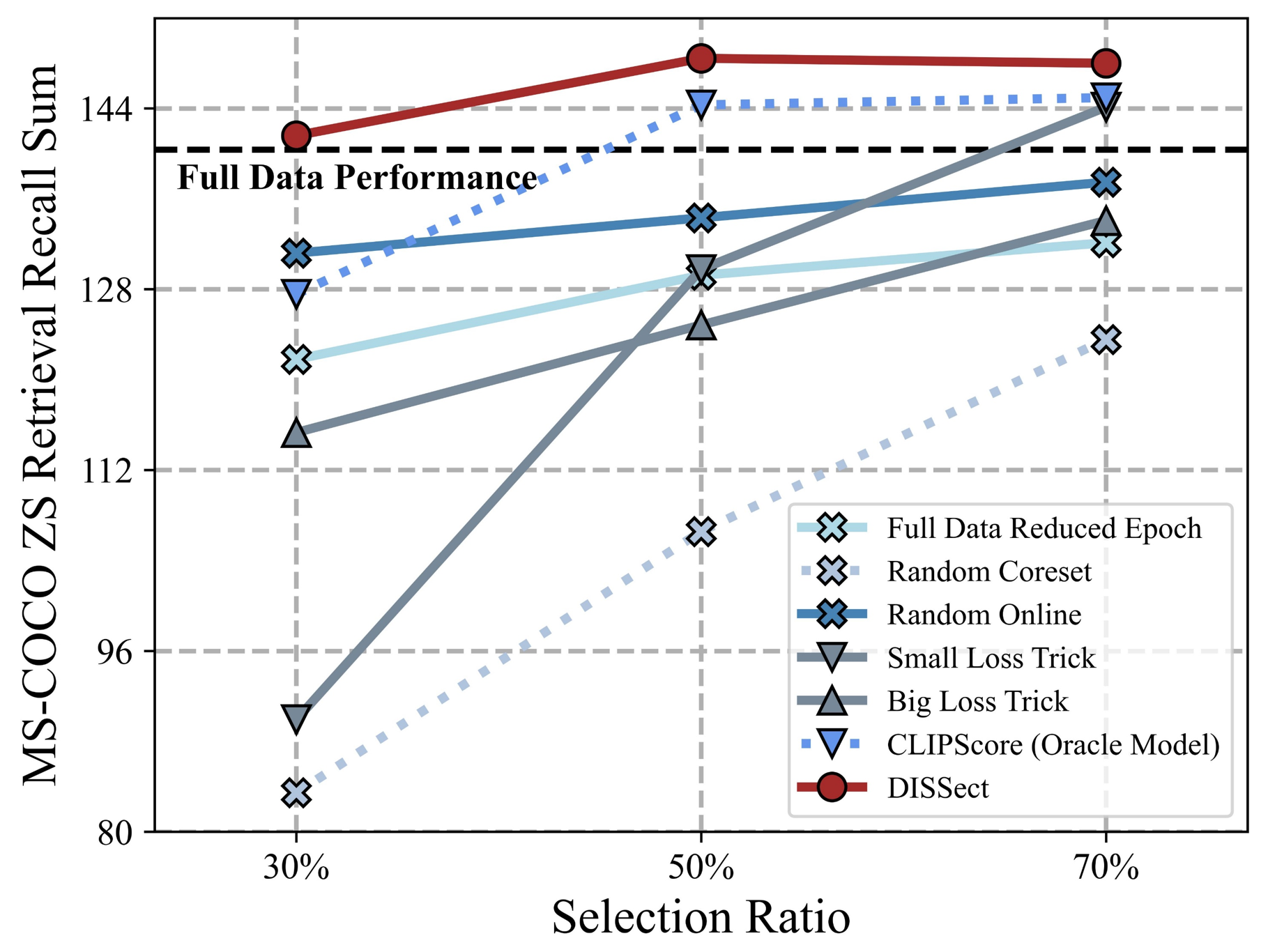}
   \caption{Performance of various sample selection methods under different selection ratios pre-trained on CC3M dataset and tested on MS-COCO zero-shot retrieval task. Online and coreset selection methods are separately marked with full and dotted lines. }
   \label{fig:1_method}
   \vspace{-7pt}
\end{figure}

While sample selection in unimodal applications typically focuses on filtering out simple or redundant samples~\citep{qin2023infobatch, abbas2023semdedup}, moderate selection strategies remain under-explored in multimodal scenarios. To identify the key factors that can accelerate multimodal contrastive learning, we conduct a heuristic experiment comparing different selection baselines, as shown in Fig.~\ref{fig:1_method}. The results lead to several key conclusions. 

\noindent
$\circ$ \textit{\textbf{Online selection methods generally outperform coreset selection methods.}} From the optimization aspect, a similar strategy under online selection paradigm (\textit{i.e.} Random Online) can outperform that under coreset selection (\textit{i.e.} Random Coreset) due to their greater data accessibility, which is also supported by previous researches \citep{qin2023infobatch, hong2024diversified}.

\noindent
$\circ$ \textit{\textbf{Loss-scale-based selection strategies under-perform in multimodal contrastive learning.}} Traditional loss-scale-based selection in unimodal settings, such as the ``big loss" and ``small loss" tricks (which select samples with either larger or smaller losses) aims to sample informative or clean data during training. However, in the context of contrastive learning, both methods can perform worse than random selection due to the biased judgment by the model prediction.

\noindent
$\circ$ \textit{\textbf{Precisely excluding noisy correspondence data benefits contrastive learning with high efficiency and better performance.}} In comparison, CLIPScore~\citep{hessel2021clipscore} achieves an overall superior performance than full data training with less computational cost, far surpassing the ``small loss" trick which also aims at filtering out low-quality data. CLIPScore reveals the model's judgment on the similarity of the given data pair, which can be expressed as follows,
\begin{equation}
\begin{aligned}
\text{CLIPScore}(I_i,T_i) = w*\max{(f(I_i)^\top g(T_i), 0)}
 \label{equ:1_method}
 \end{aligned}
\end{equation}
\noindent
where $w$ is a constant to control the scale for better visualization and we adopt $w = 100$ in this study. However, the success of CLIPScore is due to leveraging a CLIP-L model pre-trained on the large-scale Laion-400M dataset~\citep{schuhmann2021laion}. We refer to it as an \textbf{oracle model} that provides authoritative assessments. The filtered samples by CLIPScore typically share lower similarity scores, indicating that these samples are weakly matched or mismatched, \textit{i.e.}, they are samples with \textbf{noisy correspondence}. On the contrary, performance of the ``small loss" trick rapidly declines under low selection ratios, demonstrating its weakness to correctly identify noisy correspondence data.

\begin{figure*}[t]
  \centering
   \includegraphics[width=1.0\linewidth]{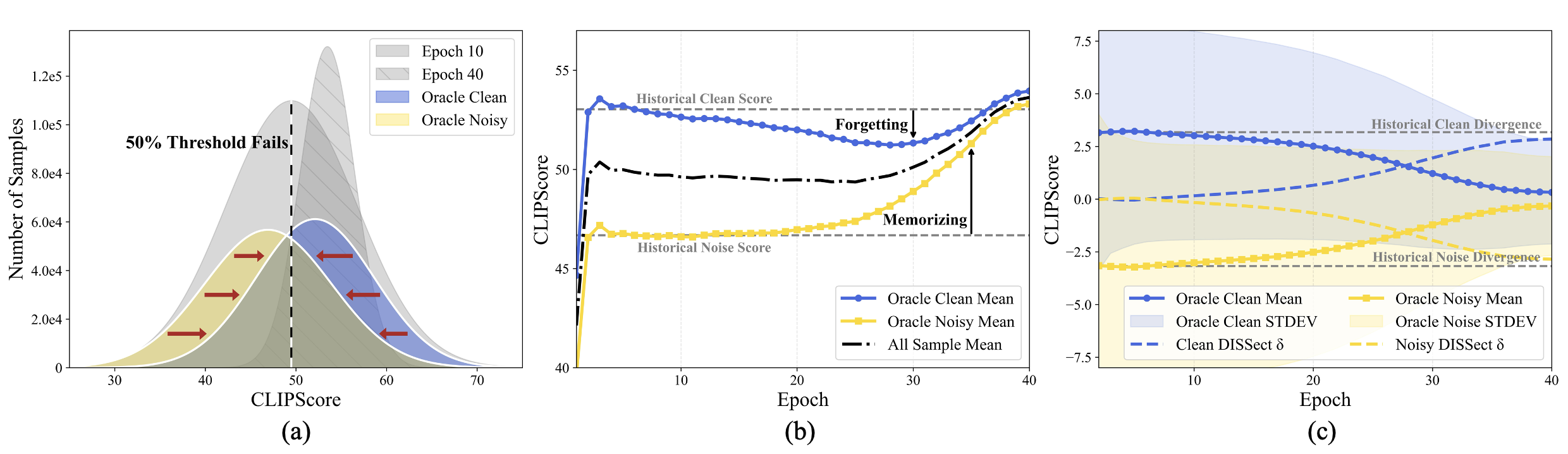}
   \caption{Memorization effect on contrastive learning. (a) CLIPScore distributions at different training stages, with clean and noisy distributions separated by oracle model. (b) CLIPScore changing curves over epochs, illustrating different learning tendencies of clean and noisy samples. (c) CLIPScore changing divergence excluding mean value fluctuations. All experiments are conducted on CC3M dataset.}
   \label{fig:2_method}
   \vspace{-7pt}
\end{figure*}

\subsection{Re-examining the Memorization Effect}

While CLIPScore offers an effective data-cleaning mechanism, obtaining an oracle model is not always feasible in the cold-start scenarios. Furthermore, relying merely on clean but simple samples (\textit{i.e.,} small-loss samples) may not be optimal throughout the entire pre-training process as verified in the previous section. In this paper, we focus on designing an \textit{oracle-model-free} selection method that tailors batched samples based solely on the predictions of the under-training model, while still preventing learning from noisy correspondences, much like an oracle model would. 

First, let us re-examine the progress of experiments in Fig.~\ref{fig:1_method}. In Fig.~\ref{fig:2_method}(a), we record the CLIPScore distributions in the $10$th and $40$th epochs during training of the vanilla model on the full dataset. As can be seen, noisy correspondence directly impacts the divergence of the learned cross-modal similarity, with the variance of CLIPScore significantly shrinking during the late epochs. This attributes to the memorization effect~\citep{liu2020early, wei2024memorization} which characterizes the tendency of deep learning models that first learn the clean patterns within the dataset before memorizing the noisy patterns. However, when we divide the training data into relatively cleaner and noisier partitions by the oracle model, we can find some interesting clues. Through Fig.~\ref{fig:2_method}(a), we can find that there is always a growing overlap between clean and noisy data similarity distributions. This explains the failure of the ``small loss" trick, making it impossible to precisely separate them using a simple loss threshold~\citep{guo2024scan} or a Gaussian Mixture Model~\citep{li2020dividemix}. To combat with this issue, a naive strategy is early-stopping~\citep{yao2007early}, which halts model training in an earlier stage before it memorizes the noisy data. However, this approach can also ``halt" the performance gain, as the model fails to fully converge. 

Besides, from the perspective of the CLIPScore mean, we can better analyze the learning trend. As shown in Fig.~\ref{fig:2_method}(b), the model quickly learns the similarities between positive pairs in the first epoch, with clean data initially yielding higher mean CLIPScore than the noisy ones. However, as training progresses, the model begins to forget the high similarity of clean data, with CLIPScore slightly dropping from its earlier peak. In contrast, the noisy correspondences are gradually memorized causing a rapid increase in CLIPScore, ultimately resulting in the collapse together. 

\subsection{Differential-informed Sample Selection}

We can better visualize the phenomenon of memorization effect when excluding the fluctuations in the mean value of CLIPScore, as shown in Fig.~\ref{fig:2_method}(c), where the relative CLIPScore of both clean and noisy data approach each other with training proceeding. As it becomes increasingly difficult to distinguish between clean and noisy data based on the current model predictions, we conclude that learning from noisy correspondence involves a trade-off that sacrifices the model’s discernibility. To combat, we propose the \underline{D}ifferential-\underline{I}nformed \underline{S}ample \underline{S}el\underline{ect}ion (DISSect), which quantifies this sacrifice by calculating the accumulated differential $\delta$ between the CLIPScore of the historical model state and that of the current model, as following expressed,
\begin{equation}
\begin{aligned}
\delta = \text{CLIPScore}_{\text{hist}} - \text{CLIPScore}_{\text{curr}}
 \label{equ:2_method}
 \end{aligned}
\end{equation}

Practically, the historical status can be obtained from an early-learning point, \textit{i.e.}, predicting $\text{CLIPScore}_{\text{hist}}$ after a few epochs of warm-up. This differential effectively distinguishes between samples that are forgotten or wrongly memorized. As shown in Fig.~\ref{fig:2_method}(c), for noisy samples, $\text{CLIPScore}_{\text{curr}}$ typically shows a significant increase compared to $\text{CLIPScore}_{\text{hist}}$, which distinctly differs from clean data pairs. Unlike the overlapping trend in CLIPScore distributions, the proposed differential $\delta$ highlights a clear and growing gap between clean and noisy correspondences. Given our goal of emphasizing forgotten samples while eliminating learning from noisy correspondences, we select samples with the greatest $\delta$ during training, \textit{i.e.}, $\mathcal{D}_b'(f,g) = \text{topk}_r\delta$, where $r$ is the selection ratio budget.

\noindent
\textbf{Theoretical Insights.} We further give theoretical insights into the principle behind DISSect’s ability to discriminate noisy correspondence samples. Based on the theorem in \citep{liu2020early}, the memorization effect arises from the influence of noisy data on gradient descent. Since the majority of data pairs are correctly matched, the gradient descent $-\nabla \mathcal{L}$ is well-aligned with the correct direction until the early-learning point, when most clean samples are adequately learned and their gradients approach zero. Meanwhile, noisy data remains unmemorized and exhibit large gradients. As training progresses, the over-parameterized model begins to memorize the noisy patterns, eventually causing the gradients of all samples to approach zero. By calculating the partial derivative of the contrastive loss in Eq.~\ref{equ:0_mthod} with respect to $\big<f(I_i), g(T_i)\big>$, we demonstrate in detail in the Appendix that during the later stages of training,
\begin{equation}
\begin{aligned}
\frac{\partial{\mathcal{L}}}{\partial{\big<f(I_i),g(T_i)\big>}} \propto \frac{1}{|\mathcal{D}_b|}\sum_{i=1}^{|\mathcal{D}_b|}\exp{(\text{CLIPScore})}^{-1}
 \label{equ:3_method}
 \end{aligned}
\end{equation}

\noindent
which indicates that the proposed differential in CLIPScore actually reflects the gradient offset of samples across different training stages. This differential captures the decrease in gradient values caused by model memorization of noisy data, in contrast to the minor changes in the gradients of clean data after the early learning point.

\noindent
\textbf{Adaptive estimation of $\text{CLIPScore}_{\text{hist}}$.} Since the warm-up strategy requires prior specification of the early-learning epoch, we also provide an adaptive estimation of the historical status $\text{CLIPScore}_{\text{hist}}$ with temporal ensembling~\citep{laine2016temporal}. The historical status of epoch $t$ can be updated by the momentum function as follows,
\begin{equation}
\small
\begin{aligned}
 \text{CLIPScore}_{\text{hist}}^{[t]} = \beta \text{CLIPScore}_{\text{hist}}^{[t-1]} + (1-\beta) \text{CLIPScore}_{\text{curr}}^{[t]}
 \label{Equal: Momentum Update}
 \end{aligned}
\end{equation}

\noindent
where $\beta\in (0,1)$ is the momentum param.  Through experiments, the temporal ensembling strategy demonstrates comparable performance to the warm-up approach, while the model remains relatively insensitive to the hyper-parameter $\beta$. The training procedure of warm-up version can refer to the pseudo code~\ref{algorithm:0_method}. The training procedure of the temporal ensembling version can be referred in the Appendix.


\noindent
\textbf{Discussion.} Our proposed DISSect is founded on the principle of eliminating noisy correspondences within data batches.  Inspired by empirical observations, we efficiently achieve this through a simple yet effective accumulated differential of CLIPScore. Compared to existing sample selection methods, our method requires no oracle model and delivers significant acceleration. Unlike noisy-label learning techniques such as ELR~\citep{liu2020early}, which differs in the learning objective and the optimization process, we tackle noise correspondences issue at the data level through efficient selection, ensuring a more substantial, lossless acceleration.

\begin{algorithm}[t]
\KwIn{Dataset $\mathcal{D}$, Warm-up Epochs $T_{w}$, Selection Ratio $r$.}
{
\For{each epoch $t = 1,2,\ldots,T_w$}{
    Learn Encoders $f_{\theta_1}$, $g_{\theta_2}$ by loss Eq.~\eqref{equ:0_mthod}.
}

Predict $\text{CLIPScore}_{\text{hist}}$ by Eq.~\eqref{equ:1_method}.

\For{each epoch $t = T_w,\ldots,T$}{
\For{each batch $\mathcal{D}_b$ from $\mathcal{D}$}{
    Forward-propagation to get features $\{f(I),g(T)\}$.
    
    Predict $\text{CLIPScore}_{\text{curr}}$ by Eq.~\eqref{equ:1_method}.
    
    Compute discrepancy score $\delta$ by Eq.~\eqref{equ:2_method}.

    Extract mini-batch $\mathcal{D}_b'$ from $\mathcal{D}_b$ with $\text{topk}_r \delta$.

    Calculate loss by Eq.~\eqref{equ:0_mthod} on $\mathcal{D}_b'$, back-propagation.

}    
}
}

\KwOut{Pre-training accelerated encoders $f_{\theta_1}$, $g_{\theta_2}$.}
\caption{Pipeline of learning with DISSect.}
\label{algorithm:0_method}
\end{algorithm}
\section{Experiments}
\subsection{Experimental Setup}
\noindent
\textbf{Datasets.} We conduct vision-language pre-training on multiple datasets, including CC3M~\citep{sharma2018conceptual}, YFCC15M~\citep{thomee2016yfcc100m} and CC12M~\citep{changpinyo2021conceptual}. To verify the overall performance and generalization ability, we further evaluate on comprehensive down-stream tasks, including image-caption retrieval~\citep{karpathy2015deep}, image reasoning~\citep{li2019visual} and image captioning~\citep{antol2015vqa}, using datasets including MS-COCO~\citep{lin2014microsoft}, Flickr30K~\citep{young2014image}, and NLVR$^2$~\citep{suhr2018corpus}. Further dataset details can be found in the Appendix.

\noindent
\textbf{Implementation details.} We implement our method based on PyTorch~\citep{paszke2019pytorch}, which is further trained on 8 NVIDIA A100 GPUs with 80GB Memory. The model framework is chosen as the CLIP-ResNet50 during experiments unless specified otherwise. We train it with a batch size of 1024 on every GPU for a total of 40 epochs on the CC3M dataset and 20 epochs on YFCC15M and CC12M datasets. For fair comparison, we follow the default setting recommended by OpenCLIP~\citep{cherti2023reproducible, ilharco_gabriel_2021_5143773}, including the AdamW~\citep{loshchilov2017decoupled} optimizer with a cosine annealing learning rate scheduler. For inputs, images are randomly cropped to a resolution of 224. The evaluation code of down-stream tasks is based on CLIP-Benchmark~\citep{clip_benchmark} for CLIP models provided by LAION-AI and BLIP evaluation code~\citep{li2022blip} for BLIP architecture. 

\noindent
\textbf{Baselines.} In addition to training on full data and random sampling, we further compare our method with various baseline methods, including coreset selection methods like forgetting score~\citep{toneva2018empirical}, CLIPScore~\cite{hessel2021clipscore}, CLIPLoss~\citep{wang2025cliploss}, and online batch selection methods like CLIPScore*, the small loss trick, big loss trick, DivBS~\citep{hong2024diversified}, InfoBatch~\citep{qin2023infobatch} and SCAN~\citep{guo2024scan}, where CLIPScore* is specifically implemented as the online selection version of CLIPScore. CLIPScore, CLIPLoss and SCAN are methods originally designed for multimodal settings. Notably, CLIPScore and CLIPLoss both introduce oracle models for reference, while our method only relies on the self model prediction.

\begin{table*}[!t]
\caption{Comparison on zero-shot retrieval performance of different selection methods pre-trained on the CC3M dataset under 30\%, 50\% and 70\% selection ratios. The upper and lower panel separately records the results on the 5K MS-COCO and 1K Flickr30K test sets. IR@K and TR@K separately represent the Recall@K score of image and text retrieval. Full data training results are colored in \textcolor{gray}{gray} and the best selection results are marked by \textbf{bold}. Notably, both CLIPScore and CLIPLoss have adopted an oracle model for reference.}
\vspace{-8pt}
\label{1_table}
\begin{center}
\begin{small}
\resizebox{2\columnwidth}{!}{%
\begin{tabular}{c|c|cccc|cccc|cccc}
\toprule[1.5pt]
\multicolumn{2}{c|}{ZS MS-COCO (5K Test Set)}          & \multicolumn{4}{c|}{30\% Selection Ratio}    & \multicolumn{4}{c|}{50\% Selection Ratio}    & \multicolumn{4}{c}{70\% Selection Ratio}    \\
\multicolumn{2}{c|}{Methods}                  & IR@1 & IR@10 & TR@1 & TR@10 & IR@1 & IR@10 & TR@1 & TR@10 & IR@1 & IR@10 & TR@1 & TR@10 \\
\midrule[0.6pt]
\textcolor{gray}{Full Data}                       & \textcolor{gray}{CLIP}~\citep{cherti2023reproducible}       &      \multicolumn{4}{c}{}     &    \textcolor{gray}{21.42}   &    \textcolor{gray}{56.92}  &   \textcolor{gray}{15.38}    &  \multicolumn{1}{c}{\textcolor{gray}{46.65}}   & \multicolumn{4}{c}{}       \\
\midrule[0.6pt]
 & Random     & 10.32 & 36.48 & 7.66  & 28.99 & 14.00 & 45.42 & 10.57 & 36.55 & 17.54 & 51.36 & 12.89 & 41.77 \\
                         Coreset       & Forgetting~\citep{toneva2018empirical}  & 12.18 & 40.08 & 8.74  & 32.78 & 14.92 & 45.86 & 10.92 & 37.65 & 17.76 & 51.92 & 12.60 & 41.63 \\
                        Selection        & CLIPScore~\citep{hessel2021clipscore}       & 18.88 & 52.50 & 13.51 & 42.65 & 22.54 & 58.14 & 15.88 & 47.28 & \textbf{23.74} & 59.74 & 16.51 & 48.36 \\
                                & CLIPLoss~\citep{wang2025cliploss}   & 19.26 & 53.02 & 13.41 & 42.53 & 21.86 & 58.26 & 15.75 & 46.97 & 23.40 & \textbf{60.00} & 16.66 & 48.41 \\
\midrule[0.6pt]
 \multirow{7}{*}{\shortstack{Online\\ Selection}}& Random    & 19.36 & 53.50 & 14.08 & 44.24 & 19.70 & 55.14 & 14.33 & 45.13 & 21.08 & 56.20 & 14.56 & 45.64 \\
 & Big Loss   & 15.54 & 48.50 & 11.56 & 39.70 & 18.14 & 51.74 & 13.06 & 41.91 & 20.04 & 55.46 & 14.17 & 44.42 \\
                                 & Small Loss  & 11.74 & 38.08 & 8.70  & 31.40 & 20.05 & 53.64 & 13.26 & 42.81 & 22.18 & 57.68 & 16.42 & 47.85 \\
                               & CLIPScore*~\citep{hessel2021clipscore} & 11.08 & 40.00 & 8.90  & 33.48 & 17.96 & 50.54 & 13.22 & 42.01 & 20.82 & 56.18 & 15.24 & 45.71 \\
                                & DivBS~\citep{hong2024diversified}      & 19.26 & 53.70 & 13.93 & 43.93 & 18.76 & 53.79 & 14.19 & 45.25 & 20.21 & 54.56 & 14.90 & 45.64 \\
                                & InfoBatch~\citep{qin2023infobatch}  & 19.60 & 54.98 & 13.80 & 44.80 & 20.58 & 54.92 & 14.36 & 45.22 & 20.96 & 56.68 & 15.28 & 45.77 \\
                                & SCAN~\citep{guo2024scan}      & 19.30 & 53.90 & 14.20 & 44.92 & 20.64 & 56.40 & 15.27 & 46.54 & 21.44 & 56.40 & 15.42 & 47.09 \\
\midrule[0.6pt] 
\cellcolor[HTML]{EFEFEF}& \cellcolor[HTML]{EFEFEF}DISSect-Mome.   & \cellcolor[HTML]{EFEFEF}20.62 & \cellcolor[HTML]{EFEFEF}55.20 & \cellcolor[HTML]{EFEFEF}15.63 & \cellcolor[HTML]{EFEFEF}47.63 & \cellcolor[HTML]{EFEFEF}22.60 & \cellcolor[HTML]{EFEFEF}57.94 & \cellcolor[HTML]{EFEFEF}\textbf{16.96} & \cellcolor[HTML]{EFEFEF}\textbf{49.81} & \cellcolor[HTML]{EFEFEF}22.46 & \cellcolor[HTML]{EFEFEF}58.66 & \cellcolor[HTML]{EFEFEF}\textbf{16.71} & \cellcolor[HTML]{EFEFEF}49.38 \\
\multirow{-2}{*}{\cellcolor[HTML]{EFEFEF}Ours}                                 & \cellcolor[HTML]{EFEFEF}DISSect-Warmup     & \cellcolor[HTML]{EFEFEF}\textbf{21.34} & \cellcolor[HTML]{EFEFEF}\textbf{56.40} & \cellcolor[HTML]{EFEFEF}\textbf{15.90} & \cellcolor[HTML]{EFEFEF}\textbf{47.95} & \cellcolor[HTML]{EFEFEF}\textbf{23.58} & \cellcolor[HTML]{EFEFEF}\textbf{58.44} & \cellcolor[HTML]{EFEFEF}16.76 & \cellcolor[HTML]{EFEFEF}49.66 & \cellcolor[HTML]{EFEFEF}22.74 & \cellcolor[HTML]{EFEFEF}58.92 & \cellcolor[HTML]{EFEFEF}16.65 & \cellcolor[HTML]{EFEFEF}\textbf{49.70} \\
\bottomrule[1.5pt]
\toprule[1.5pt]
\multicolumn{2}{c|}{ZS Flickr30K (1K Test Set)}          & \multicolumn{4}{c|}{30\% Selection Ratio}    & \multicolumn{4}{c|}{50\% Selection Ratio}    & \multicolumn{4}{c}{70\% Selection Ratio}    \\
\multicolumn{2}{c|}{Methods}                  & IR@1 & IR@10 & TR@1 & TR@10 & IR@1 & IR@10 & TR@1 & TR@10 & IR@1 & IR@10 & TR@1 & TR@10 \\
\midrule[0.6pt]
\textcolor{gray}{Full Data}                       & \textcolor{gray}{CLIP}~\citep{cherti2023reproducible}       &      \multicolumn{4}{c}{}     &    \textcolor{gray}{39.10}   &    \textcolor{gray}{78.30}  &   \textcolor{gray}{29.38}    &  \multicolumn{1}{c}{\textcolor{gray}{67.66}}   & \multicolumn{4}{c}{}       \\
\midrule[0.6pt]
 & Random     & 17.00 & 52.50 & 14.30 & 43.06 & 29.80 & 66.50 & 21.42 & 55.46 & 35.40 & 74.40 & 25.54 & 62.06 \\
                    Coreset            & Forgetting~\citep{toneva2018empirical} & 21.30 & 59.30 & 16.46 & 21.30 & 28.00 & 68.40 & 21.20 & 55.34 & 31.40 & 72.40 & 24.60 & 60.90 \\
                    Selection            & CLIPScore~\citep{hessel2021clipscore}  & 33.40 & 74.90 & 25.50 & 62.54 & 39.50 & 77.00 & 30.18 & 65.98 & 41.70 & \textbf{81.30} & 31.68 & 69.02 \\
                                & CLIPLoss~\citep{wang2025cliploss}   & 34.10 & 72.30 & 26.48 & 60.68 & 39.60 & 77.60 & 29.38 & 66.20 & \textbf{43.00} & 81.10 & 32.12 & 68.74 \\
\midrule[0.6pt]
  \multirow{7}{*}{\shortstack{Online\\ Selection}}& Random     & 35.10 & 76.20 & 28.90 & 65.34 & 39.50 & 76.90 & 28.12 & 66.56 & 38.80 & 76.20 & 28.50 & 64.72 \\
  & Big Loss   & 28.00 & 72.10 & 21.64 & 58.98 & 33.90 & 74.30 & 25.92 & 61.90 & 37.10 & 75.10 & 27.44 & 63.12 \\
                                & Small Loss & 25.00 & 57.40 & 16.50 & 47.34 & 33.90 & 73.40 & 27.18 & 61.86 & 38.50 & 78.60 & 30.96 & 67.22 \\
                                & CLIPScore*~\citep{hessel2021clipscore} & 21.90 & 55.40 & 16.26 & 46.68 & 31.70 & 68.40 & 23.94 & 59.16 & 40.10 & 76.60 & 28.30 & 64.74 \\
                                & DivBS~\citep{hong2024diversified}      & 35.60 & 73.00 & 26.72 & 64.92 & 36.20 & 77.20 & 27.52 & 66.78 & 36.60 & 77.20 & 28.20 & 67.20 \\
                                & InfoBatch~\citep{qin2023infobatch}  & 36.90 & 76.20 & 26.82 & 65.20 & 37.00 & 75.30 & 28.20 & 65.12 & 37.80 & 76.50 & 29.14 & 66.90 \\
                                & SCAN~\citep{guo2024scan}       & 36.50 & 75.40 & 26.70 & 64.80 & 38.30 & 76.20 & 28.68 & 65.92 & 38.50 & 78.40 & 29.68 & 66.98 \\
\midrule[0.6pt]        
\cellcolor[HTML]{EFEFEF}& \cellcolor[HTML]{EFEFEF}DISSect-Mome.   & \cellcolor[HTML]{EFEFEF}37.10 & \cellcolor[HTML]{EFEFEF}\textbf{76.70} & \cellcolor[HTML]{EFEFEF}29.80 & \cellcolor[HTML]{EFEFEF}\textbf{68.01} & \cellcolor[HTML]{EFEFEF}\textbf{42.70} & \cellcolor[HTML]{EFEFEF}78.10 & \cellcolor[HTML]{EFEFEF}30.92 & \cellcolor[HTML]{EFEFEF}\textbf{70.28} & \cellcolor[HTML]{EFEFEF}39.00 & \cellcolor[HTML]{EFEFEF}78.80 & \cellcolor[HTML]{EFEFEF}\textbf{32.73} & \cellcolor[HTML]{EFEFEF}69.20 \\
 \cellcolor[HTML]{EFEFEF}\multirow{-2}{*}{Ours}          & \cellcolor[HTML]{EFEFEF}DISSect-Warmup     & \cellcolor[HTML]{EFEFEF}\textbf{40.50} & \cellcolor[HTML]{EFEFEF}76.00 & \cellcolor[HTML]{EFEFEF}\textbf{30.40} & \cellcolor[HTML]{EFEFEF}67.72 & \cellcolor[HTML]{EFEFEF}39.40 & \cellcolor[HTML]{EFEFEF}\textbf{79.20} & \cellcolor[HTML]{EFEFEF}\textbf{32.52} & \cellcolor[HTML]{EFEFEF}69.98 & \cellcolor[HTML]{EFEFEF}40.60 & \cellcolor[HTML]{EFEFEF}80.00 &\cellcolor[HTML]{EFEFEF} 32.58 & \cellcolor[HTML]{EFEFEF}\textbf{70.70} \\
\bottomrule[1.5pt]
\end{tabular}}
\vspace{-8pt}
\end{small}
\end{center}
\end{table*}

\subsection{Experimental Results}

\noindent
\textbf{Image-caption retrieval pre-trained on CC3M.} The performance of down-stream image-caption retrieval tasks directly reflects the model’s ability to accurately identify true positive cross-modal data pairs. In Table~\ref{1_table}, we pre-train the CLIP model using various selection methods on the CC3M dataset and subsequently evaluate the models on down-stream MS-COCO and Flickr30K image-caption retrieval tasks for comparison. At lower selection ratios, online selection methods generally outperform coreset selection methods due to their greater data accessibility, with random online selection serving as a strong baseline since its alignment with the original dataset distribution. At higher selection ratios, CLIPScore and CLIPLoss outperform random online selection by leveraging an oracle model, underscoring the importance of eliminating noisy correspondence samples. Our method DISSect achieves the best performance across all selection ratios, surpassing most baseline methods including those relying on oracle models, demonstrating its effectiveness. Notably, DISSect performs comparably to full data training with a 30\% selection ratio. Additionally, the warm-up version DISSect-Warmup slightly outperforms temporal ensembling version DISSect-Mome., while the latter requires fewer hyper-parameters.

\begin{table}[!t]
\caption{Comparison on zero-shot retrieval performance pre-trained on CC12M and YFCC15M. Full data training results are colored in \textcolor{gray}{gray} and best selection results are marked by \textbf{bold}.}
\vspace{-8pt}
\label{2_table}
\begin{center}
\begin{small}
\resizebox{1.0\columnwidth}{!}{%
\begin{tabular}{c|c|ccccc}
\toprule[1.5pt]
Datasets                 & Methods   & IR@1  & IR@10 & TR@1  & TR@10 & R@Sum  \\
\midrule[0.6pt]
\multicolumn{7}{c}{Train: 50\% Selection Ratio, Eval: MS-COCO (5K Test Set)} \\
\midrule[0.6pt]
\multirow{4}{*}{CC12M}   & \textcolor{gray}{Full Data} & \textcolor{gray}{33.40} & \textcolor{gray}{71.88} & \textcolor{gray}{23.49} & \textcolor{gray}{59.39} & \textcolor{gray}{188.16} \\
                         & Random    & 32.56 & 69.60 & 20.52 & 56.11 & 178.79 \\
                         & SCAN      & 33.50 & 71.04 & 22.10 & 58.10 & 184.74 \\
                         & \cellcolor[HTML]{EFEFEF}DISSect      & \cellcolor[HTML]{EFEFEF}\textbf{34.10} & \cellcolor[HTML]{EFEFEF}\textbf{71.94} & \cellcolor[HTML]{EFEFEF}\textbf{22.52} & \cellcolor[HTML]{EFEFEF}\textbf{59.05} & \cellcolor[HTML]{EFEFEF}\textbf{187.61} \\
\midrule[0.6pt]
\multirow{4}{*}{YFCC15M} & \textcolor{gray}{Full Data} & \textcolor{gray}{28.24} & \textcolor{gray}{65.42} & \textcolor{gray}{18.03} & \textcolor{gray}{50.74} & \textcolor{gray}{162.43} \\
                         & Random    & 29.04 & 66.44 & 17.98 & 51.52 & 164.98 \\
                         & SCAN      & 28.06 & 65.10 & 17.27 & 50.66 & 161.09 \\
                         & \cellcolor[HTML]{EFEFEF}DISSect      & \cellcolor[HTML]{EFEFEF}\textbf{30.16} & \cellcolor[HTML]{EFEFEF}\textbf{67.80} & \cellcolor[HTML]{EFEFEF}\textbf{18.05} & \cellcolor[HTML]{EFEFEF}\textbf{51.78} & \cellcolor[HTML]{EFEFEF}\textbf{167.79} \\
\midrule[0.6pt]
\multicolumn{7}{c}{Train: 50\% Selection Ratio, Eval: Flickr30K (1K Test Set)} \\
\midrule[0.6pt]
\multirow{4}{*}{CC12M}   & \textcolor{gray}{Full Data} & \textcolor{gray}{57.50} & \textcolor{gray}{89.70} & \textcolor{gray}{44.28} & \textcolor{gray}{81.58} & \textcolor{gray}{273.06} \\
                         & Random    & 54.30 & 86.70 & 40.66 & 78.00 & 259.66 \\
                         & SCAN      & 55.40 & 87.00 & 42.68 & 79.38 & 264.46 \\
                         & \cellcolor[HTML]{EFEFEF}DISSect      & \cellcolor[HTML]{EFEFEF}\textbf{56.30} & \cellcolor[HTML]{EFEFEF}\textbf{87.10} & \cellcolor[HTML]{EFEFEF}\textbf{43.94} & \cellcolor[HTML]{EFEFEF}\textbf{80.18} & \cellcolor[HTML]{EFEFEF}\textbf{267.52} \\
\midrule[0.6pt]
\multirow{4}{*}{YFCC15M} & \textcolor{gray}{Full Data} & \textcolor{gray}{46.30} & \textcolor{gray}{86.10} & \textcolor{gray}{30.26} & \textcolor{gray}{67.04} & \textcolor{gray}{229.70} \\
                         & Random    & 48.00 & 85.50 & 31.48 & 68.72 & 233.70 \\
                         & SCAN      & 49.10 & 86.90 & 31.90 & 70.98 & 238.88 \\
                         & \cellcolor[HTML]{EFEFEF}DISSect      & \cellcolor[HTML]{EFEFEF}\textbf{50.90} & \cellcolor[HTML]{EFEFEF}\textbf{87.50} & \cellcolor[HTML]{EFEFEF}\textbf{33.82} & \cellcolor[HTML]{EFEFEF}\textbf{71.46} & \cellcolor[HTML]{EFEFEF}\textbf{243.68} \\
\bottomrule[1.5pt]
\end{tabular}}
\vspace{-8pt}
\end{small}
\end{center}
\end{table}

\noindent
\textbf{Image-caption retrieval pre-trained on larger datasets.} To demonstrate the efficiency of our method in larger-scale scenarios, we further pre-train the CLIP model under 50\% selection ratio on the CC12M and YFCC15M datasets. CC12M consists of 12 million well-curated, high-quality image-text pairs, collected by relaxing the data collection pipeline in a manner similar to CC3M. YFCC15M is a 15 million subset of the large-scale multilingual YFCC100M dataset, containing noisier English captions. The results on downstream retrieval tasks are reported in Table~\ref{2_table}. In both datasets, DISSect outperforms random sampling and the SCAN method. Notably, DISSect achieves significantly better performance on the noisier YFCC15M dataset, far surpassing full data training. These results highlight the performance gain of eliminating noisy correspondence and effectiveness of DISSect for handling noisy, real-world data.

\begin{table}[!t]
\caption{Comparison on different down-stream tasks of selection methods pre-trained on CC3M. Full data training results are colored in \textcolor{gray}{gray} and best selection results are marked by \textbf{bold}.}
\vspace{-8pt}
\label{table: cc3m retrieval}
\begin{center}
\begin{small}
\resizebox{1\columnwidth}{!}{%
\begin{tabular}{c|cc|cccc}
\toprule[1.5pt]
\multirow{2}{*}{Methods} & \multicolumn{2}{c|}{NLVR$^2$ Reasoning} & \multicolumn{4}{c}{COCO Captioning} \\
                         & dev        & test-P      & B@4              & METEOR              & CIDEr              & SPICE            \\
\midrule[0.6pt]
\multicolumn{7}{c}{Train: 50\% Selection Ratio on BLIP-Base} \\
\midrule[0.6pt]
\textcolor{gray}{Full Data}                & \textcolor{gray}{76.09}      & \textcolor{gray}{76.44}       & \textcolor{gray}{37.11}             & \textcolor{gray}{29.49}             & \textcolor{gray}{123.36}            & \textcolor{gray}{22.52}           \\
Random                   &     68.46       &     68.85        & 36.04             & 28.58             & 119.65            & 21.74           \\
SCAN                   &     72.19       &     72.59        & 36.04             & 29.02             & 120.44            & 22.07           \\
\cellcolor[HTML]{EFEFEF}DISSect                     &    \cellcolor[HTML]{EFEFEF}\textbf{73.99}        &       \cellcolor[HTML]{EFEFEF}\textbf{75.07}      & \cellcolor[HTML]{EFEFEF}\textbf{37.16}             & \cellcolor[HTML]{EFEFEF}\textbf{29.37}             & \cellcolor[HTML]{EFEFEF}\textbf{123.91}            & \cellcolor[HTML]{EFEFEF}\textbf{22.46}           \\
\bottomrule[1.5pt]
\end{tabular}}
\vspace{-8pt}
\end{small}
\end{center}
\end{table}

\noindent
\textbf{Comparisons on other down-stream tasks.} We further evaluate our method on other downstream tasks, including visual reasoning tasks on the NLVR$^2$ dataset and image captioning tasks on the COCO dataset. The Natural Language Visual Reasoning (NLVR$^2$) task is a binary classification task that involves reasoning about a question in relation to two given images. The image captioning task involves generating a description for an input image. Both tasks assess the cross-modal understanding ability of the multimodal model. We apply different selection methods to the BLIP~\citep{li2022blip} backbone, which is with generalization ability through optimizing on captioning loss. DISSect outperforms the SCAN method on both downstream tasks, with comparable performance on the COCO captioning task to the full data training, indicating the effectiveness.

\begin{table}[!t]
\caption{Ablation on implementation with CLIP/ResNet101 and BLIP/ViT-B architectures. Full data training results are colored in \textcolor{gray}{gray} and best selection results are marked by \textbf{bold}.}
\vspace{-8pt}
\label{4_table}
\begin{center}
\begin{small}
\resizebox{1.0\columnwidth}{!}{%
\begin{tabular}{c|c|ccccc}
\toprule[1.5pt]
Backbones                 & Methods   & IR@1  & IR@10 & TR@1  & TR@10 & R@Sum  \\
\midrule[0.6pt]
\multicolumn{7}{c}{Train: 50\% Selection Ratio, Eval: MS-COCO (5K Test Set)} \\
\midrule[0.6pt]
      & \textcolor{gray}{Full Data} & \textcolor{gray}{23.20} & \textcolor{gray}{60.22} & \textcolor{gray}{17.05} & \textcolor{gray}{49.55} & \textcolor{gray}{150.02} \\
                CLIP/            & Random    & 20.74 & 57.26 & 15.12 & 47.00 & 140.12 \\
                ResNet101            & SCAN      & 21.56 & 57.92 & 15.85 & 48.19 & 143.52 \\
                            & \cellcolor[HTML]{EFEFEF}DISSect      & \cellcolor[HTML]{EFEFEF}\textbf{23.54} & \cellcolor[HTML]{EFEFEF}\textbf{59.86} & \cellcolor[HTML]{EFEFEF}\textbf{16.87} & \cellcolor[HTML]{EFEFEF}\textbf{49.91} & \cellcolor[HTML]{EFEFEF}\textbf{150.18} \\
\midrule[0.6pt]
 & \textcolor{gray}{Full Data} & \textcolor{gray}{38.48} & \textcolor{gray}{73.74} & \textcolor{gray}{30.15} & \textcolor{gray}{64.83} & \textcolor{gray}{207.20} \\
            BLIP/                & Random    & 36.88 & 71.20 & 28.98 & 63.70 & 200.76 \\
            ViT-B                & SCAN      & 37.44 & 72.24 & 28.86 & 63.70 & 202.24 \\
                            & \cellcolor[HTML]{EFEFEF}DISSect      & \cellcolor[HTML]{EFEFEF}\textbf{38.42} & \cellcolor[HTML]{EFEFEF}\textbf{73.72} & \cellcolor[HTML]{EFEFEF}\textbf{29.84} & \cellcolor[HTML]{EFEFEF}\textbf{64.40} & \cellcolor[HTML]{EFEFEF}\textbf{206.38} \\
\midrule[0.6pt]
\multicolumn{7}{c}{Train: 50\% Selection Ratio, Eval: Flickr30K (1K Test Set)} \\
\midrule[0.6pt]
      & \textcolor{gray}{Full Data} & \textcolor{gray}{43.30} & \textcolor{gray}{82.10} & \textcolor{gray}{33.26} & \textcolor{gray}{72.30} & \textcolor{gray}{230.96} \\
            CLIP/                & Random    & 40.50 & 78.40 & 29.24 & 67.56 & 215.70 \\
            ResNet101                & SCAN      & 40.60 & 79.50 & 30.04 & 68.48 & 218.62 \\
                            & \cellcolor[HTML]{EFEFEF}DISSect      & \cellcolor[HTML]{EFEFEF}\textbf{41.00} & \cellcolor[HTML]{EFEFEF}\textbf{79.40} & \cellcolor[HTML]{EFEFEF}\textbf{32.30} & \cellcolor[HTML]{EFEFEF}\textbf{71.16} & \cellcolor[HTML]{EFEFEF}\textbf{223.86} \\
\midrule[0.6pt]
 & \textcolor{gray}{Full Data} & \textcolor{gray}{69.80} & \textcolor{gray}{94.20} & \textcolor{gray}{54.62} & \textcolor{gray}{85.10} & \textcolor{gray}{303.72} \\
            BLIP/                & Random    & 62.20 & 92.20 & 50.18 & 84.56 & 289.14 \\
            ViT-B                & SCAN      & 63.30 & 92.60 & 50.92 & 84.42 & 291.24 \\
                            & \cellcolor[HTML]{EFEFEF}DISSect      & \cellcolor[HTML]{EFEFEF}\textbf{65.70} & \cellcolor[HTML]{EFEFEF}\textbf{93.80} & \cellcolor[HTML]{EFEFEF}\textbf{51.26} & \cellcolor[HTML]{EFEFEF}\textbf{84.68} & \cellcolor[HTML]{EFEFEF}\textbf{295.44} \\
\bottomrule[1.5pt]
\end{tabular}}
\vspace{-8pt}
\end{small}
\end{center}
\end{table}

\begin{table}[!t]
\caption{Comparison on \textbf{zero-shot} retrieval performance of our method to different noisy correspondence strategies pre-trained on the CC3M dataset. The best results are marked by \textbf{bold}.}
\vspace{-8pt}
\label{5_table}
\begin{center}
\begin{small}
\resizebox{1.0\columnwidth}{!}{%
\begin{tabular}{c|c|ccccc}
\toprule[1.5pt]
Methods    & Dual Net. & IR@1  & IR@10 & TR@1  & TR@10 & R@Sum  \\
\midrule[0.6pt]
\multicolumn{7}{c}{MS-COCO (5K Test Set)} \\
\midrule[0.6pt]
Small Loss &     $\times$        & 20.05 & 53.64 & 13.26 & 42.81 & 129.76 \\
NCR~\citep{huang2021learning}        &     \checkmark         & 20.90 & 56.14 & 15.16 & 45.79 & 137.99 \\
GSC~\citep{zhao2024mitigating}        &       \checkmark       & 22.26 & 57.56 & 16.21 & 48.69 & 144.72 \\
\cellcolor[HTML]{EFEFEF}DISSect       &      \cellcolor[HTML]{EFEFEF}$\times$        & \cellcolor[HTML]{EFEFEF}\textbf{23.58} & \cellcolor[HTML]{EFEFEF}\textbf{58.44} & \cellcolor[HTML]{EFEFEF}\textbf{16.76} & \cellcolor[HTML]{EFEFEF}\textbf{49.66} & \cellcolor[HTML]{EFEFEF}\textbf{148.44} \\
\midrule[0.6pt]
\multicolumn{7}{c}{Flickr30K (1K Test Set)} \\
\midrule[0.6pt]
Small Loss &       $\times$       & 33.90 & 73.40 & 27.18 & 61.86 & 196.34 \\
NCR~\citep{huang2021learning}        &      \checkmark        & \textbf{40.00} & 77.20 & 30.46 & 68.28 & 215.94 \\
GSC~\citep{zhao2024mitigating}        &       \checkmark       & 39.50 & \textbf{80.40} & 29.18 & 67.30 & 216.38 \\
\cellcolor[HTML]{EFEFEF}DISSect       &     \cellcolor[HTML]{EFEFEF}$\times$         & \cellcolor[HTML]{EFEFEF}39.40 & \cellcolor[HTML]{EFEFEF}79.20 & \cellcolor[HTML]{EFEFEF}\textbf{32.52} & \cellcolor[HTML]{EFEFEF}\textbf{69.98} & \cellcolor[HTML]{EFEFEF}\textbf{221.10} \\
\bottomrule[1.5pt]
\end{tabular}}
\vspace{-8pt}
\end{small}
\end{center}
\end{table}

\subsection{Experimental Analysis}

\begin{figure*}[!t]
  \centering
   \includegraphics[width=1.0\linewidth]{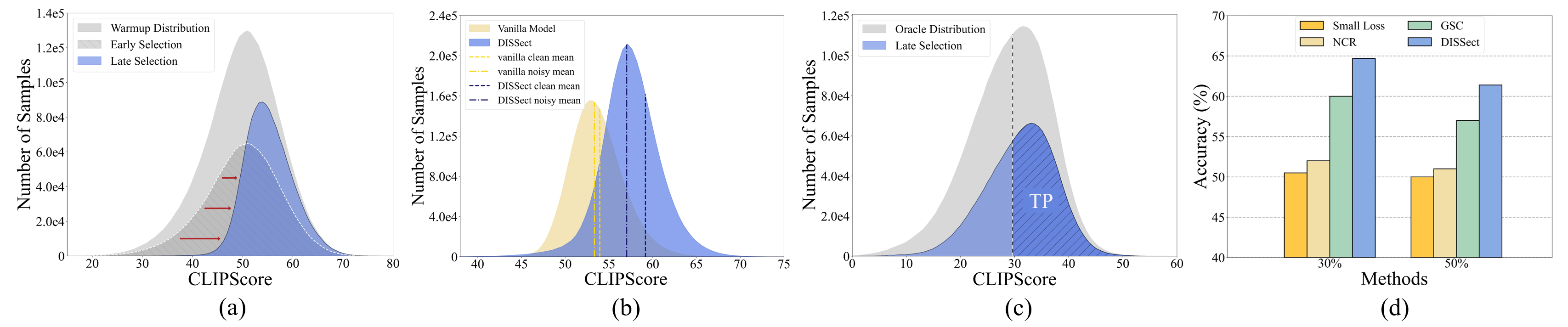}
   \caption{Selection strategy analysis on CC3M dataset. (a) CLIPScore distribution of selected data by DISSect at different training stages with respect to all samples. (b) DISSect learns a higher and more diverse CLIPScore distribution compared to the vanilla model. (c) Through comparison with the oracle model demonstrates that DISSect effectively prevents learning from noisy correspondence. (d) DISSect achieves higher True Positive accuracy than NCR and GSC on discriminating samples with noisy correspondence.}
   \label{fig:3_experiment}
   \vspace{-6pt}
\end{figure*}

\noindent
\textbf{Selection strategy analysis.} To investigate the practical sampling strategy of DISSect, we visualize the changing tendency of CLIPScore distribution on selected samples with regard to all samples in Fig.~\ref{fig:3_experiment}(a). Rather than relying solely on clean samples throughout the training procedure, DISSect begins with a random sampling approach during the early epochs, while gradually shifting toward cleaner samples in the later epochs. This strategy benefits multimodal learning by improving generalization, as it allows the model to access a broader range of data before memorizing noisy correspondence samples. In Fig.\ref{fig:3_experiment}(b), we visualize the final CLIPScore distribution learned by DISSect compared to the vanilla model. DISSect learns a higher and more diverse similarity distribution, with the discrepancy between clean and noisy data increasing from 0.65 to 2.14. 

\begin{figure}[!t]
  \centering
   \includegraphics[width=1.0\linewidth]{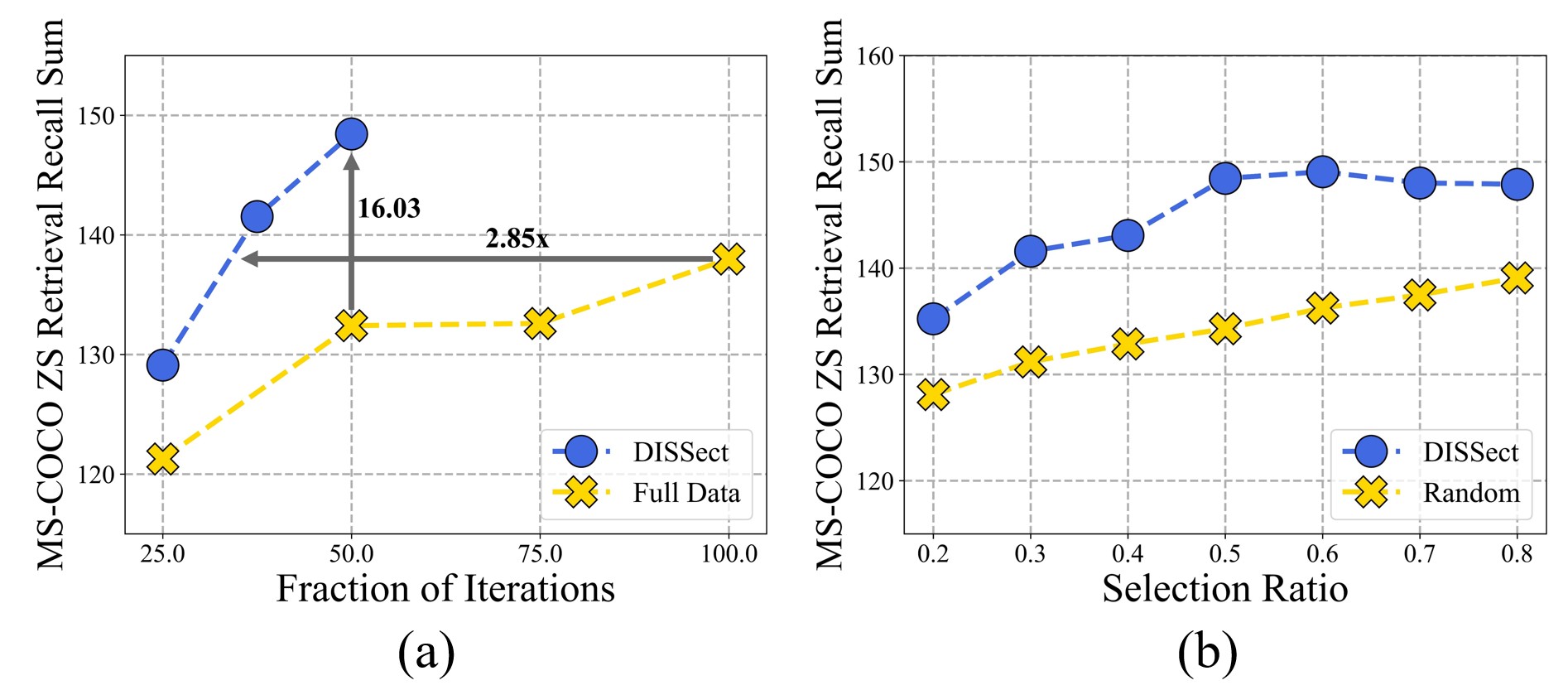}
   \caption{Ablation and analysis. (a) Comparison on training efficiency to full data training on MS-COCO retrieval tasks. (b) Ablation study on various selection ratios, compared to random sampling. Both experiments are pre-trained on CC3M dataset.}
   \label{fig:4_experiment}
   \vspace{-8pt}
\end{figure}

\noindent
\textbf{Comparison to noisy correspondence methods.} To assess DISSect’s ability to discriminate noisy correspondence samples, we introduce an evaluation metric based on the oracle model. Specifically, as shown in Fig.\ref{fig:3_experiment}(c), we compare the CLIPScore distribution of selected samples to the oracle model’s distribution, which can be further segmented by a 50\% threshold to compute the True Positive (TP) accuracy. DISSect achieves an accuracy of 65\% under a 30\% selection ratio and 61\% under a 50\% selection ratio. These accuracy results surpass the “small loss” trick and modern noisy correspondence methods, including NCR~\citep{huang2021learning} and GSC~\citep{zhao2024mitigating}, as shown in Fig.\ref{fig:3_experiment}(d). Notably, both NCR and GSC utilize a dual-network design in a co-teaching manner to enhance performance, which however requires more than twice the computational cost than full data training. In Tab.~\ref{5_table}, we compare DISSect to noisy correspondence methods on the down-stream retrieval tasks, where DISSect achieves the best performance with a single network architecture.

\noindent
\textbf{Adaptation analysis.} Our proposed method is easy to implement and plug-and-play. In Tab.~\ref{4_table}, we apply our method to both CLIP/ResNet101 and BLIP/ViT-B. DISSect relies solely on the image-text alignment (ITA) module to select samples when applied to the BLIP backbone. After applied to such larger models, the overall performance on downstream tasks shows significant improvement, with DISSect consistently outperforming both random sampling and the SCAN method across both backbone architectures.

\noindent
\textbf{Acceleration analysis.} As illustrated in Fig.~\ref{fig:4_experiment}(a), DISSect achieves approximately 2.85$\times$ acceleration to reach the same performance as full data training, and outperforms full data training by 16.03 in Recall Sum with the same number of iteration steps. The computation time for convergence is recorded in Tab.~\ref{6_table}. With the exception of uniform sampling, online batch selection methods generally require an additional forward pass for each batch. This forward propagation can be further accelerated by utilizing low-precision cores~\citep{jouppi2017datacenter} or employing a group of workers for asynchronous execution on sample selection~\citep{alain2015variance}. The overhead of both versions of DISSect method is negligible.

\begin{table}[!t]
\caption{Wall-clock computation time ($\downarrow$) of different methods training on single GPU on CC3M and CC12M datasets. ``To Rand. Best" refers to the time cost to reach same performance as random sampling, as well as ``To DISSect Best". The subscript indicates the percentage of time saved compared to full data training.}
\vspace{-8pt}
\label{6_table}
\begin{center}
\begin{small}
\resizebox{1.0\columnwidth}{!}{%
\begin{tabular}{c|c|c|c c}
\toprule[1.5pt]
Full Data & Selection & Random & \multicolumn{2}{c}{\cellcolor[HTML]{EFEFEF}DISSect} \\
Training & Ratio &  \footnotesize To Rand. Best  & \cellcolor[HTML]{EFEFEF}\footnotesize To Rand. Best & \cellcolor[HTML]{EFEFEF}\footnotesize To DISSect Best \\
\midrule[0.6pt]
\multicolumn{5}{c}{CC3M Dataset (40 epochs) } \\
\midrule[0.6pt]
\multirow{3}{*}{58.3 h}  & 30\%            &   17.8 h $_{\downarrow 69\%}$  & \cellcolor[HTML]{EFEFEF}15.3 h $_{\downarrow 74\%}$ &  \cellcolor[HTML]{EFEFEF}21.8 h $_{\downarrow 63\%}$   \\
                   & 50\%            &    30.2 h $_{\downarrow 49\%}$    & \cellcolor[HTML]{EFEFEF}21.8 h $_{\downarrow 63\%}$ &   \cellcolor[HTML]{EFEFEF}36.4 h $_{\downarrow 38\%}$  \\
                   & 70\%            &   41.1 h $_{\downarrow 29\%}$    &  \cellcolor[HTML]{EFEFEF}29.3 h $_{\downarrow 50\%}$ &  \cellcolor[HTML]{EFEFEF}50.9 h $_{\downarrow 13\%}$   \\
\midrule[0.6pt]
\multicolumn{5}{c}{CC12M Dataset (20 epochs)} \\
\midrule[0.6pt]
\multirow{3}{*}{116.1 h}  & 30\%            &   35.5 h $_{\downarrow 69\%}$  & \cellcolor[HTML]{EFEFEF}32.5 h $_{\downarrow 72\%}$ &  \cellcolor[HTML]{EFEFEF}43.4 h $_{\downarrow 63\%}$   \\
                   & 50\%            &    60.2 h $_{\downarrow 49\%}$   & \cellcolor[HTML]{EFEFEF}50.7 h $_{\downarrow 56\%}$ &  \cellcolor[HTML]{EFEFEF}72.5 h $_{\downarrow 38\%}$  \\
                   & 70\%            &  82.5 h $_{\downarrow 29\%}$     &  \cellcolor[HTML]{EFEFEF}62.3 h $_{\downarrow 46\%}$  &  \cellcolor[HTML]{EFEFEF}101.3 h  $_{\downarrow 13\%}$   \\
\bottomrule[1.5pt]
\end{tabular}}
\vspace{-8pt}
\end{small}
\end{center}
\end{table}

\noindent
\textbf{Ablation study on selection ratios.} In Fig.~\ref{fig:4_experiment}(b), we record the performance of DISSect across different selection ratios. DISSect consistently outperforms random sampling across all selection ratios. DISSect maintains stable performance at higher selection ratios ($>$50\%) and performs comparably to full data training even at 30\% selection ratio. The best performance is observed at 60\% selection ratio.
\section{Conclusion}
In this paper, we propose the novel online selection method Differential-informed Sample Selection (DISSect) to accelerate multimodal training. Specifically, we highlight the critical role of eliminating noisy correspondence on contrastive learning and propose that the differential between the predicted correlation of the current model and that of a historical model reflects the cleanliness samples. Extensive experiments  demonstrate the consistent effectiveness and efficiency of DISSect over state-of-the-art methods.

\noindent
\textbf{Acknowledgement.} This work is supported by the National Key R\&D Program of China (No.2022ZD0160702) and National Natural Science Foundation of China (No.62306178).
{
    \small
    \bibliographystyle{ieeenat_fullname}
    \bibliography{main}
}


\end{document}